\documentclass{article}
\pdfoutput=1

\usepackage[utf8]{inputenc}
\usepackage{graphicx}
\usepackage{color,colortbl}
\usepackage[svgnames]{xcolor}
\usepackage{booktabs}
\usepackage{enumitem}
\usepackage{array}
\usepackage{url}
\usepackage[hidelinks,colorlinks=true,allcolors=MediumBlue]{hyperref}
\usepackage[authoryear,sort&compress,square]{natbib}
\usepackage{amsmath}
\usepackage{makecell}
\usepackage{amssymb}
\usepackage{multirow}
\usepackage{multicol}
\usepackage{microtype}
\usepackage{abstract}
\usepackage[export]{adjustbox}
\usepackage{layouts}
\usepackage{fontawesome}
\usepackage{titling}

\usepackage[bottom=30pt,top=80pt,left=124pt,bottom=80pt,right=124pt,headsep=0pt,footskip=30pt]{geometry}

\usepackage[scaled]{helvet}
\usepackage[T1]{fontenc}
\usepackage{sectsty}
\allsectionsfont{\sffamily}

\makeatletter\newcommand{\subtitle}[1]{\gdef\@subtitle{#1}}
\newcommand{\@subtitle}{}

\def\@maketitle{%
  \newpage
  \null
  \vspace{-2em}
  \sffamily
  \begin{center}%
  \let \footnote \thanks
    {\Huge\bfseries\@title \par}
    \vskip 1em%
    {\LARGE\mdseries\@subtitle \par}
    \vskip 2em%
    {\large \@author}
    \vskip 1em%
  \end{center}%
  \par
  \vskip 1.5em}
\makeatother

\let\cite\citep

\newcommand{\dri}[1]{}

\newcommand{\gemma}[1]{Gemma-#1}
\newcommand{\llama}[1]{LLaMA #1}
\newcommand{\qwen}[1]{Qwen-#1}

\newcommand{\gpt}[1]{GPT-#1}

\title{Apple Intelligence Foundation Language Models}
\subtitle{Tech Report 2025}
\author{Apple}
\date{July 2025}

\begin{document}
\renewcommand{\abstractname}{\vspace{-20pt}}

\maketitle
\hypersetup{pdfauthor={Apple},pdftitle={Apple Intelligence Foundation Language Models: Tech Report 2025}}

\begin{abstract}
\sffamily
\noindent
We introduce two multilingual, multimodal foundation language models that power Apple Intelligence features across Apple devices and services:
(i) a $\sim$3B-parameter on-device model optimized for Apple silicon through architectural innovations such as KV-cache sharing and 2-bit quantization-aware training;
and (ii) a scalable server model built on a novel Parallel-Track Mixture-of-Experts (PT-MoE) transformer that combines track parallelism, mixture-of-experts sparse computation, and interleaved global–local attention to deliver high quality with competitive cost on Apple's Private Cloud Compute platform.
Both models are trained on large-scale multilingual and multimodal datasets sourced via responsible web crawling, licensed corpora, and high-quality synthetic data, then further refined with supervised fine-tuning and reinforcement learning on a new asynchronous platform.
The resulting models support several additional languages while understanding images and executing tool calls.
In public benchmarks and human evaluations, both the server model and the on-device model match or surpass comparably sized open baselines.

A new Swift-centric Foundation Models framework exposes guided generation, constrained tool calling, and LoRA adapter fine-tuning, allowing developers to integrate these capabilities with a few lines of code. The latest advancements in Apple Intelligence models are grounded in our Responsible AI approach with safeguards like content filtering and locale-specific evaluation, as well as our commitment to protecting our users' privacy with innovations like Private Cloud Compute.
\end{abstract}

\section{Introduction \dri{Ruoming}}

    Apple Intelligence integrates powerful generative AI right into the apps and experiences users turn to everyday, all while protecting their privacy. At the 2025 Worldwide Developers Conference we introduced a wide range of new Apple Intelligence features that help users communicate, express themselves, and achieve their goals. To power these features, we have created a new generation of language foundation models specifically developed to enhance the Apple Intelligence features in our latest software releases. We also introduced the new Foundation Models framework, which gives app developers direct access to the on-device language foundation model at the core of Apple Intelligence.

    We crafted these generative models to power the wide range of intelligent features integrated across our platforms. The models have improved tool-use and reasoning capabilities, understand image and text inputs, are faster and more efficient, and are designed to support 16 different languages. Our latest foundation models include a compact, approximately 3-billion-parameter model optimized to run efficiently on Apple silicon, alongside a mixture-of-expert server-based model with a novel architecture tailored for Private Cloud Compute~\cite{ApplePCC}. These two foundation models are part of a larger family of generative models created by Apple to support our users.

    In this overview, we detail the data we used for training, the architectures of the models we designed, the training recipes we employed, the techniques we used to optimize inference, and our evaluation results when compared to similar models. Throughout, we highlight how we achieved an expansion of capabilities and quality improvements for the benefits of our users while also increasing the speed and efficiency of on-device and Private Cloud Compute. Finally, in our continued commitment to uphold our core values, we illustrate how Responsible AI principles are integrated throughout the entire model development process.

\begin{figure}[!ht]
    \includegraphics[width=1.2\textwidth,alt={Modeling Overview Diagram.},center]{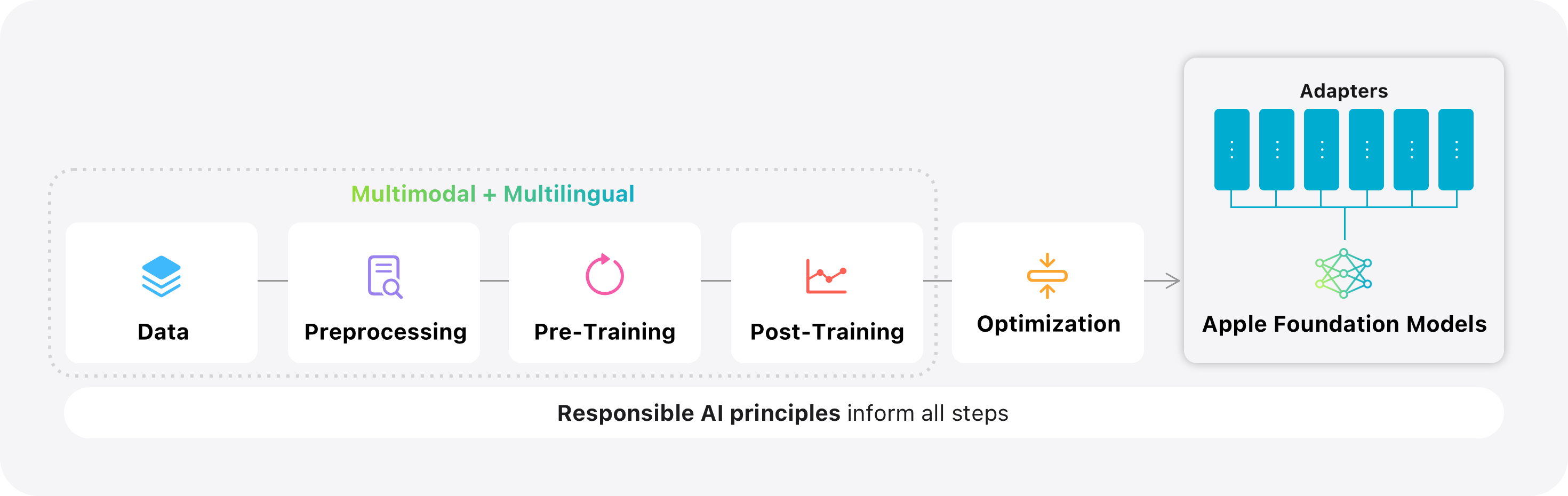}
    \vspace{-1.5em}
    \caption{Modeling overview for the Apple foundation models.}
    \label{fig:overview}
\end{figure}

\section{Model Architectures}
\label{sec:modelarch}

We developed both the on-device and server models to meet a wide range of performance and deployment needs. The on-device model is optimized for efficiency, enabling low-latency inference with minimal resource usage, while the server model is designed to deliver high accuracy and scalability for more complex tasks. Together, they form a complementary suite of solutions adaptable to diverse application scenarios. We first describe the general decoder architectures for both models, then explain how we extended them to support image inputs with vision encoders. 

\subsection{On-Device Model}
We incorporated the following improvements into the on-device model, resulting in a significant reduction in inference latency and improved efficiency.

\paragraph{KV Cache Sharing.} To improve the efficiency of our on-device architecture, we divided the full model into two blocks. Block 1 contained 62.5\% of the total transformer layers, while Block 2 contained the remaining 37.5\% of the transformer layers, but had the key and value projections removed. Every key and value cache (KV cache) of Block 2 was directly shared with those generated by Block 1~\cite{SwiftKV}, reducing KV cache memory usage by 37.5\%. Additionally, because Block 2 does not produce any keys or values, the prefill stage is able to bypass all of its computation~\cite{sun2024you} and reduce time-to-first-token (TTFT) by $\sim$37.5\%.
    
\subsection{Server Model}
We also made several major changes to improve the capacity and efficiency of the server model.
    
\paragraph{Parallel Track (PT) Transformer.} 
We introduced a new architecture, the Parallel Track (PT) Transformer. Unlike the standard decoder-only transformer, which consists of a single sequential stack of layers, the PT-transformer partitions the model into multiple smaller transformers, referred to as tracks. Each track consists of multiple stacked track blocks, each of which is a Transformer layer stack. The track blocks process tokens independently, with synchronization across tracks applied only at the input and output boundaries of each track block. This isolated design enables straightforward parallel execution across tracks and reduces the synchronization overhead found in conventional transformer decoders, such as those using tensor parallelism. We refer to this approach as track parallelism. This improved training and inference latency without compromising model quality.
    
\paragraph{Parallel Track Mixture of Experts (PT-MoE).} 
To further scale the server-side model, we introduced Mixture-of-Experts (MoE) layers~\cite{noam2017, du2022, gshard, zoph2022} within each track block, leading to the PT-MoE architecture. Since the experts in each MoE layer are local to their respective tracks, the communication overhead can be more effectively overlapped with the computation to improve training efficiency. Combined with track-level independence enabled by track parallelism, this design allowed the model to scale efficiently while maintaining low latency due to increased sparsity. 

\begin{figure}
    \includegraphics[width=1.2\linewidth,alt={PT-MoE Diagram.},center]{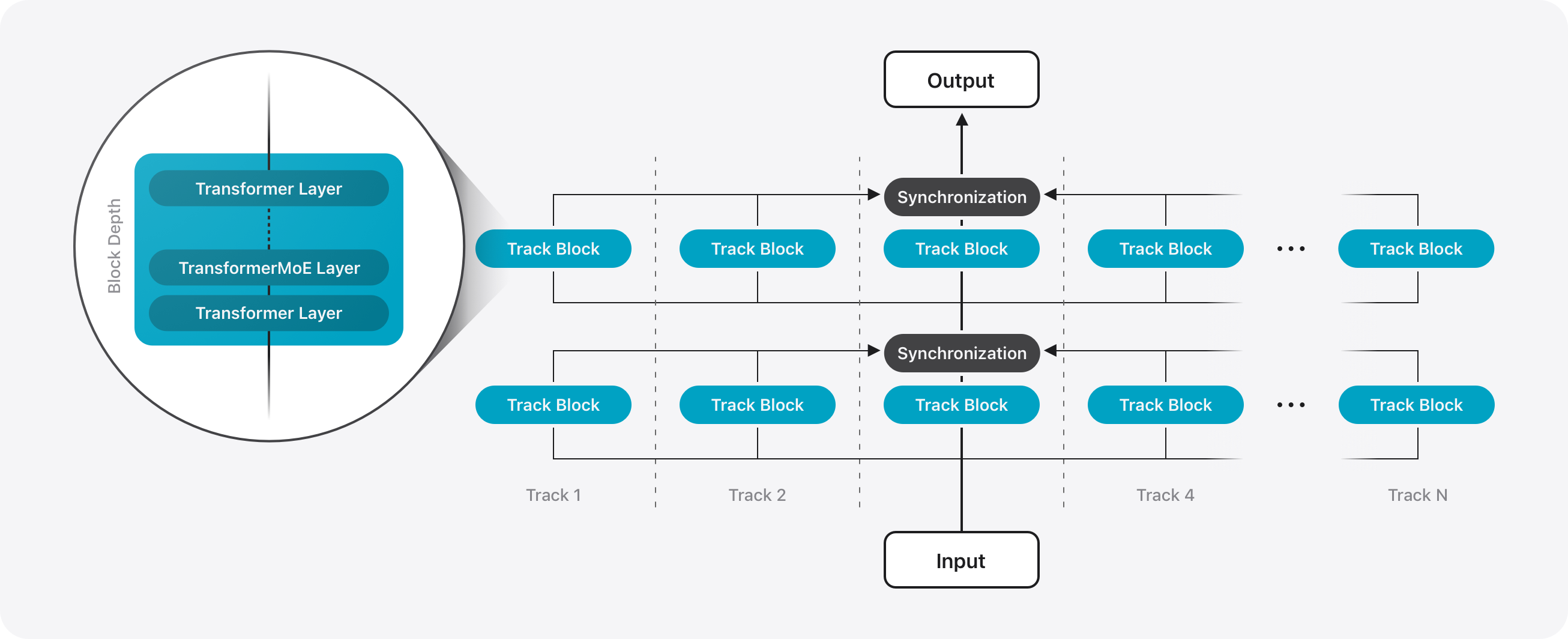}
    \vspace{-1.5em}
    \caption{Diagram of the PT-MoE architecture. Each track is composed of multiple track blocks, and each track block contains a fixed number of transformer/MoE layers. Assume that we have a total of $L$ layers and track block of depth $D$, then we reduce the synchronization overhead from $2L$ (tensor parallelism) to $L/D$ (track parallelism). For example, if $D=4$, the PT reduces $87.5\%$ of the synchronization overhead.}
    \label{fig:pt-moe}
\end{figure}

Specifically, we replaced the dense feed-forward network in every other transformer layer with a MoE layer where the top-k routing is implemented via the grouped general matrix multiplication (GEMM)~\cite{megablox}, ensuring that no tokens are dropped during routing. \autoref{fig:pt-moe} provides an overview of the PT-MoE architecture.

\paragraph{Interleaving Global and Local Attention Layers.} 
    We designed an interleaved attention architecture that alternates between local and global attention layers to support long sequences efficiently~\cite{Beltagy2020Longformer}. Each repeating transformer block consists of three local attention layers using a sliding window of size 4096 and rotational positional embeddings (RoPE), followed by a global attention layer without positional embeddings (NoPE)~\cite{yang2025ropenopeagainnew}. Omitting positional embeddings in the global attention layer allows better length generalization by avoiding out-of-distribution position issues in long contexts. This interleaved design maintains the model's quality while substantially reducing the KV cache size for long-context inference.

\subsection{Vision Encoder \dri{Yanghao Li (v8 draft ready)} }\label{sec:image_encoder}
To enable visual understanding capabilities, we introduced a visual encoder that can extract vision features from input images. The visual encoder was pre-trained on a large amount of image data to enhance its performance. 

The vision encoder includes two key components: a vision backbone to extract rich visual representations from input images, and a vision-language adaptation module to compress the vision representations and align these visual features with the token representation of the model.

\paragraph{Vision Backbone.} For the vision backbone, we adopted the standard Vision Transformer (ViT-g)~\cite{dosovitskiy2020image} with 1B parameters for the server model and the more efficient ViTDet-L backbone~\cite{li2022exploring} with 300M parameters for the on-device model.

Specifically, the on-device vision backbone adopted ViTDet, which uses window attention in most vision transformer layers with only three cross-window global attention layers. To further effectively capture and integrate both fine-grained local details and broader global contextual information, we adopted a novel Register-Window (RW) mechanism to the standard ViTDet. This approach is designed to encode a global register (or class) token by enabling it to interact with distinct local windows of an image before contributing to the overall global context aggregation. 

\paragraph{Vision-Language Adaptation Module.} The vision-language adaptation seeks to compress visual features into a fixed number of image tokens, matching the token dimension of the model. To achieve this, we employed a combination of a transformer layer, a linear projection layer, and a 3x3 convolutional layer, enabling us to capture both global and local visual information effectively. The linear projection layer specifically maps the visual features to align with the decoder dimension of the LLM. Subsequently, an average pooling layer was applied to further compress the vision features into a fixed number of image tokens.

\section{Data}
\label{sec:data}

We believe in training our models using diverse and high-quality data. This includes data that we've licensed from publishers, curated from publicly available or open-sourced datasets, and publicly available information crawled by our web-crawler, Applebot. 
 
We do not use our users’ private personal data or user interactions when training our foundation models. Additionally, we take steps to apply filters to remove certain categories of personally identifiable information and to exclude profanity and unsafe material. 
 
Further, we continue to follow best practices for ethical web crawling, including following widely-adopted robots.txt protocols to allow web publishers to opt out of their content being used to train Apple’s generative foundation models. Web publishers have fine-grained controls over which pages Applebot can see and how they are used while still appearing in search results within Siri and Spotlight.

\subsection{Web Data}

In the latest versions of our models, we have introduced several key enhancements in the pre-training data pipeline:

\label{sec:text_data}
\begin{itemize}
    \item \textbf{{Improved Web Crawling Strategy.}}  
 While respecting the opt-outs as noted above, we continue to source a significant portion of the pre-training data for our models from web content crawled by Applebot. These data span hundreds of billions of pages and cover an extensive range of languages, locales, and topics. Given the noisy nature of the web, Applebot employs advanced crawling strategies to prioritize high-quality and diverse content. In particular, we focused on capturing high-fidelity HTML pages which enrich the dataset with both text and structured metadata for aligning media with the surrounding text content. To improve relevance and quality, the system leveraged multiple signals including domain-level language identification, topic distribution analysis, and URL path pattern heuristics. 

We took special care to accurately extract the content from documents and modern websites. We enhanced our document collection with headless rendering, enabling full-page loading, dynamic content interaction, and JavaScript execution, all of which are critical for extracting data from web architectures. Special attention was given to websites that depend on dynamic content and user interactions, enabling interaction simulation and reliable extraction of meaningful information from complex pages.

    \item \textbf{Data Expansion.}  
In addition to advanced crawling strategies, we significantly expanded the scale and diversity of our training data. We incorporated a larger volume of high-quality general-domain, mathematical, and programming content. Additionally, we extended our multilingual support to new languages that will be available later this year.

    \item \textbf{Enhanced Data Extraction Pipeline.} 
    Accurate content extraction from HTML pages is critical to ensuring high-quality training data. In addition to refining extraction strategies for specific domains, we incorporated LLMs into our  extraction pipeline, particularly for domain-specific documents. LLMs have demonstrated strong performance in identifying and isolating the main content sections of documents, often outperforming traditional rule-based methods.

    \item \textbf{Refined Data Filtering.}  
    High-quality filtering plays a critical role in overall model performance. In this iteration of our models, we refined our data filtering pipeline by reducing reliance on overly aggressive heuristic rules and incorporating more model-based filtering techniques. We found that heuristics discarded valuable high-quality tokens at times. By relaxing certain rules and introducing model-informed signals, we were able to retain more informative content, resulting in a larger and higher-quality pre-training dataset. We tuned the model-based filters for each supported language by carefully selecting training documents and labels.
\end{itemize}

\subsection{Image Data \dri{Yinfei Yang (First draft done)}}

To enhance our models and enable visual understanding capabilities for Apple Intelligence features, we introduced image data into the pre-training pipeline, using high-quality licensed data along with publicly-available image data.

\subsubsection{Image-Text Crawl Data}

Using our web crawling strategy, we sourced pairs of images with corresponding alt-texts. In addition to compliance filters, we filtered for data quality, including image-text alignment. After de-duplication, this process yielded over 10B high quality image-text pairs. In addition, we created image-text interleaved data by preserving images in their originally observed text context from crawled documents. After quality and compliance filtering, this resulted in 175M interleaved image-text documents, containing over 550M images. We also use a filtered version of public image-text interleaved data, which contains 294M images.

\subsubsection{Synthetic Image Caption data}

Since web-crawled image-text pairs are generally short and often do not comprehensively describe visual details in images, we used synthetic image captioning data to provide richer descriptions. Concretely, we developed an in-house image captioning model capable of providing high quality captions at different levels of detail, ranging from a list of key words to a paragraph-level comprehensive description. We generated over 5B image-caption pairs that we used across the pre-training stages. In addition, we created a sample of particularly high quality and comprehensive synthetic captions by employing object detectors to classify the region of interest, captioning them in isolation, and then summarizing these individual sub-captions as well as the respective spatial information in the original image using an LLM to achieve even more detailed and spatial-aware image descriptions for use in later training stages.

\subsubsection{Text-Rich Image Data}
There are cases where our models need to understand images with text in them, such as to help users add an event to their calendar simply by pointing their iPhone at a flyer or printed advertisement. 
To improve our models' text-rich visual understanding capabilities, we curated various sets of text-rich data, including PDFs, documents, manuscripts, infographics, tables, and charts via licensed data, web crawling, and in-house synthesis
We then extracted the texts and generated both transcription and question-answer pairs from the image data across languages supported by Apple Intelligence.

\subsubsection{High-quality Domain-Specific Image-text Data}
To further improve our models' capabilities, we curated a variety of types of image-text data:

\begin{itemize}
    \item \textbf{High-quality caption data and grounded captions.}  
    We employed Contrastive Language-Image Pre-training (CLIP) \cite{radford2021learning} models and Optical Character Recognition (OCR) tools as filters to obtain high-quality images from the aforementioned synthetic image caption data. Then, we used an in-house grounding model to localize the nouns in the captions and append the coordinates after the nouns to form grounded captions.

    \item \textbf{Tables, charts, and plots.}  
    For charts and plots, we first prompted an internal LLM to generate synthetic data fields and corresponding values. Then we asked the LLM to write code that can generate various types of chart and plot based on the previous synthesized data samples. Lastly, we fed the charts, plots, and data samples into a teacher model to generate QAs for model training. For tables, we parsed the tables from publicly available websites and automatically converted them into markdown format, then used both the image-markdown pairs and image-synthetic QAs generated by a teacher model for model training.

    \item \textbf{Knowledge-required domains.}  
    Certain images require domain-specific knowledge to understand, such as those related to science, healthcare, and other specialized fields. To curate training data for these knowledge-required domains, we first retrieve relevant images from web-crawled datasets by matching domain-specific keywords with the alt-texts of the images. We then refine this initial selection by filtering out images with irrelevant visual content, utilizing CLIP \cite{radford2021learning} to compare the image features with the text features of category prompts. Finally we use a teacher model to synthesize questions and answers based on the retrieved images, thereby generating high-quality training data.

\end{itemize}

\section{Pre-training}

Our pre-training recipe has evolved to scale Apple Intelligence capabilities to support more languages as well as a wider array of features, including those that require image understanding.

\subsection{Text Tokenizer }
In order to better support new languages during this stage, we extended the text tokenizer from a vocabulary size of 100k to 150k, achieving representation quality for many additional languages with just 50\% more tokens.

\subsection{Vision Encoder} 
The training of the vision encoder was conducted in two stages: an initial contrastive pre-training stage followed by a joint training stage with an LLM decoder. In the first stage, we applied the CLIP method~\cite{radford2021learning} to pre-train the vision backbone using more than 6B image-text paired data, including synthetic captions as well as alt text, thereby furnishing the vision backbone with a robust initialization and strong visual grounding capabilities. For contrastive pre-training, we used a 448$\times$448 resolution. Additionally, we incorporated the masking strategy from FLIP~\cite{li2023scaling} to enhance the training efficiency of ViT-g. In the second stage, the vision backbone was jointly trained with the vision-language adaption module and a compact LLM decoder with 302M parameters to align image features with the LLM representation space. Unlike the CLIP training stage which used only image-text paired data, we enriched the training data with high-quality text data, interleaved image-text data, and specialized domain-specific image-text data. We further enhanced the vision encoder's ability to capture fine-grained details by increasing the image resolution to 672$\times$672 during this stage.

\subsection{Text Pre-training}  
We trained our server model on 8192 v5p Cloud TPU accelerators provisioned as $4 \times 2048$ chip slices using the AXLearn~\cite{axlearn} framework. Training was conducted with a combination of data parallelism, fully-sharded-data-parallel, and the newly introduced track parallelism for 13.4T tokens, where only data parallelism crosses the slice boundary. Due to built-in fault tolerance mechanisms in the AXLearn runtime we maintained an overall 93\% good output, ensuring resiliency against possible hardware or node interruption. 

Compared to last year, we made several modifications to the on-device model training pipeline to improve training efficiency and model quality. Specifically, we first trained a dense on-device model for about 14T tokens and sparse-upcycled~\cite{komatsuzaki2023sparse} it into a 64-expert, every-2-layer (i.e., interleaving dense and sparse layers) MoE~\cite{axlearnmoe} using only 1T high-quality data. Then, we retrained the dense on-device model for the last $10\%$ of tokens (about 1.4T) using a distillation loss from the MoE teacher. This new pipeline not only reduced the cost of training the large distillation teacher by $90\%$ but also greatly improved the efficiency of producing teacher logits. Moreover, retraining only the last $10\%$ of tokens rather than from scratch also eliminated the need for structural pruning. We found that this new training pipeline boosted the performance of our pretrained on-device model while drastically reducing the training costs.

\subsection{Capability Expansion through Continued Pre-training }
We adapted the pre-trained on-device and server model backbones to improve visual understanding, while also improving code, math, multilingual, and long-context understanding, using continued pre-training stages.
\begin{itemize}
    \item \textbf{Text-only continued pre-training} improved the model alignment in four key areas: Math, Code, Knowledge, and Multilingual. The training data consisted of synthetic data, our highest-quality organic data, and some fraction of the bulk pre-train data. Synthetic training examples were verified for correctness wherever possible to eliminate hallucination and included code, math, and machine-translated multilingual documents. In addition to the dataset changes, we also updated our domain mixture weights--increasing the weight allocated to code, math, and multilingual data. Specifically, we increased the total mixture weight allocated to multilingual data from 8\% to 30\%, while maintaining temperature sampling across languages within the multilingual bucket to balance the risk of underfitting and overfitting for the low-resource languages. We found that this curriculum provided the same benefit for multilingual as using a fixed 30\% weight throughout while mitigating concerns of loss of performance in English.  
    
    \item \textbf{Multimodal adaptation} allowed improvements in visual understanding, while avoiding regressions on text performance. The training data consisted of 60\% high-quality text data from the text continued pre-training stage, 10\% interleaved image-text data, 28.5\% image-text caption pairs, and 1.5\% high-quality domain-specific image-text pairs. This stage initialized the LLM decoder from the text continued pre-train stage and the image tokenizer from the image encoder training stage, while the vision-language adaptation module is trained from scratch. For the on-device model, we paired it with a RW-ViTDet visual encoder. Each image was resized to 672$\times$672 pixel resolution and represented by 144 token embeddings after being processed by the visual encoder. The on-device model was trained using 1.3T tokens at 16k sequence length. For the server model, we paired it with a ViT-g visual encoder. Each image is resized to 448$\times$448 pixel resolution and represented by 144 tokens after being processed by the visual encoder. The server model is trained using 420B tokens at 8k sequence length.
    
    \item \textbf{The context lengthening} stage increased the number of tokens the model can attend to, while maintaining its core capabilities. We accomplished this by continuing to train the model on sequences containing up to 65K tokens. These longer sequences were sampled from naturally occurring long-form data, like licensed books and code repositories, synthetic long-form data designed to target specific capabilities, such as in-context learning and retrieval, and the same data used in continued pre-training.
\end{itemize}

\section{Post-training}

Similar to pre-training, we have also evolved our post-training process to support language expansion and visual understanding. While Apple Intelligence features are powered through adapters on top of the base model, empirically we find that improving general-purpose post-training lifts the performance of all features, as the models have stronger capabilities across all use cases. To this aim, we've expanded our use of synthetic data and conducted a focused quality hill climbing on tool-use capabilities. Moreover, we also upgraded our RLHF infrastructure and recipe to include more diverse reward signals and the image modality.

\subsection{Supervised Fine-Tuning (SFT) \dri{Ke Ye (v8 draft ready)}}

We further scaled up SFT by combining human written demonstrations and synthetic data, with an emphasis on core vision capabilities. This includes: 

\begin{itemize}
  \item \textbf{General knowledge.} We collected text and image data across knowledge domains in the form of information extraction and question answering to ensure the model developed strong text and multimodal understanding abilities. We employed model-based filtering techniques to detect and improve response quality. 
  
  \item \textbf{Reasoning.} We expanded our reasoning datasets on math and coding by rewriting a large amount of web corpus, synthesizing responses and then filtering with accurate reward signals such as execution feedback, ground truth verification, and majority voting. For multi-modality, we collected several high-quality STEM datasets (including primary/junior/senior math, physics and chemistry), and also synthesized several high-quality math datasets based on publicly available datasets. Furthermore, we expanded these high-quality datasets with Chain-of-Thoughts (CoT) for sophisticated reasoning.
  
  \item \textbf{Text-rich image understanding.} We curated and synthesized image data containing rich textual information, including but not limited to handwriting, documents, infographics, charts, plots, tables, chemical formulas, and mathematical expressions. Furthermore, we augmented this dataset with data specifically curated for the parsing of documents featuring complex layout structures. Additionally, our approach involved the training of models capable of comprehending high-resolution images, enhancing their capability in processing such multifaceted data.

  \item \textbf{Multilingual Optical Character Recognition (OCR).} We incorporated high quality real-world multilingual OCR data including both printed and handwritten text such as posters, menus, store signs, notebooks, etc. We included this data to support OCR across 15 languages to be supported by Apple Intelligence.

  \item \textbf{Text and visual grounding.} We improved the quality of SFT data on grounding for tasks such as rewriting and summarization. We also expanded the visual grounding dataset to empower the model with fine-grained image understanding, grounding, and referring. This included interpreting visual prompts, such as points and bounding boxes in the form of set-of-mark and/or text annotations, and generating grounded responses by grounding text output with image bounding boxes.
  
  \item \textbf{Multi-image reasoning.} We enhanced the model's multi-image reasoning capability by incorporating additional high-quality multi-image data.
\end{itemize}
We further bootstrapped the diversity of vision SFT data through retrieval-based methods. Starting with a small set of seed prompts paired with images, we retrieved additional images through an image search pipeline and synthesized prompts and corresponding responses for those retrieved images to construct additional synthetic SFT data.
To improve robustness and mitigate hallucination, we synthesized adversarial prompts that deliberately requested information not present in the image, paired with responses that appropriately refused to answer.
We conducted ablation studies to balance the mixture ratios of each component and ensure the model was helpful, honest, and responsible. 

\paragraph{Image Resolution in SFT.} During SFT, we increased the effective resolution from $672\times 672$ pixels (default resolution processed by the vision backbone) to $1344\times 1344$ pixels by tiling images into a $2\times2$ grid.
The model then processed a total of five images, an overview image along with the four sub-images. This approach proved to be effective for high-resolution and text-rich image understanding.

To accommodate for different latency requirements, the on-device model can operate on a total of three different resolutions, with trade-offs between latency and quality: the \textit{high-resolution} mode, as mentioned above; the \textit{balanced} mode, where we provide only the overview image to the model; and a \textit{rapid} mode where the vision backbone processes an image with a resolution of $224\times 224$ pixels. In the rapid mode, the vision backbone produces only nine image tokens per image, compared to 144 in the other modes. Rapid mode is best suited to tasks that require only high-level image understanding.

To train the different resolution modes, we proceeded as follows: for low-resolution images, we flipped a coin on whether we used the rapid mode or one of the other modes, whereas for high-resolution images, we used rapid mode only for 1\% of the images. Out of the remaining images, we randomly selected 20\% of images for the balanced mode. 

\subsection{Reinforcement Learning from Human Feedback (RLHF) \dri{Dong Yin (v8 draft ready)}}

We continued to use the REINFORCE Leave-One-Out (RLOO)~\cite{ahmadian2024back} method as our main RLHF algorithm, and applied RLHF after the SFT stage for both the on-device model and the server model.

\subsubsection{Reinforcement Learning (RL) Infrastructure}

\begin{figure}[!h]
    \includegraphics[width=1.2\linewidth,alt={RL Diagram.},center]{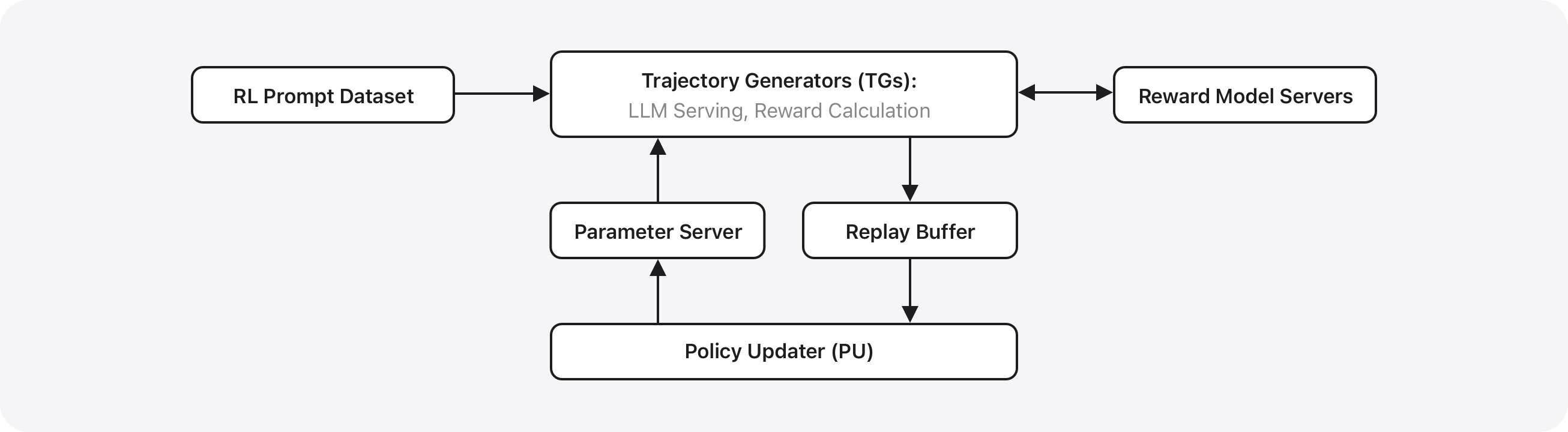}
    \vspace{-1.5em}
    \caption{Our distributed and asynchronous RL infrastructure.}
    \label{fig:rl_infra}
\end{figure}

We upgraded our RL infrastructure to be more scalable and extensible. In particular, we built a distributed asynchronous RL infrastructure consisting of replicas of trajectory generators (TGs) and a policy updater (PU). The TGs use efficient LLM serving engines to run generations given the RL prompt dataset. Given the prompt and the generated response, the TGs also calculate the reward. Our infrastructure is flexible to leverage a diverse set of reward signals, such as reward model, ground truth verification, code execution, LLM-as-a-judge, etc. The reward score can be calculated either by a remote server (such as the reward model), or by simple local computations. Once the responses and rewards are obtained, the TGs save the generated trajectories to a replay buffer. The policy updater (PU) loads the trajectories from the replay buffer and applies gradient updates to the model. The PU periodically saves the model weights to a parameter server and once the TGs detect that new model weights are saved to the parameter server, they load the new weights and use them for generation. \autoref{fig:rl_infra} illustrates our RL infrastructure described above.

There are a few benefits of our RL infrastructure. First, trajectory generation and policy improvement can happen simultaneously, and thus it is computationally more efficient than a synchronous training paradigm where the training alternates between generation and policy updates. Second, since the TGs and PU are on different devices, we can use the optimal parallelism for both of them, and use the optimized serving techniques such as continuous batching on the TGs. The inference resources can also be scaled up independently to increase the throughput. Third, as mentioned above, our design makes it more convenient to incorporate a diverse set of prompt datasets associated with different rewards in the RL training. In our experiments, we find that our new RL infrastructure can achieve similar performance compared to an earlier version of a synchronous RL system, while using $37.5\%$ less devices and $75\%$ less compute time.

\subsubsection{RLHF Recipe}

All of the parameters of the model were updated in RL training, including the vision encoder. We curated a set of prompt datasets and the associated rewards for the RLHF stage. First, we collected a large volume of text-only and image-text preference data to train a reward model. Similar to our previous model version~\cite{gunter2024apple}, we used a preference loss function along with single-sided grading as regularization to train the reward model. We then used the following prompt data categories for the RLHF stage:

\begin{itemize}
    \item Text-only prompts, including multilingual data, with rewards from the reward model.
    \item Image-text prompts, with rewards from the reward model.
    \item Math prompts, with rule-based answer verification as the reward.
    \item Image-text STEM (science, technology, engineering, and mathematics) reasoning prompts, with rule-based answer verification as the reward.
\end{itemize}

The datasets with the reward model as the reward focused on improving general human preference, such as instruction following and helpfulness, whereas the datasets with rule-based rewards focused on improving the mathematical and visual reasoning capabilities. In our experiments, we observed significant gains, especially on human preferences after applying the RLHF stage.

Since the datasets with text-only prompts, including multilingual, rely on the reward model to provide the rewards, accuracy of the reward model is critical to the effectiveness of RLHF recipe. We conducted an analysis of the consensus rate among human graders using random samples of the preference data. The analysis shows that human graders’ preferences differ for about 20-30\% of the preference data, the prompts of which are usually subjective, difficult, and/or obscure. Further analysis shows that human graders tend to have a hard time agreeing on preference labels if they likewise have difficulties assessing the overall helpfulness of the responses to the prompts. 

We trained a separate reward model, which was only used for prompt selection, with additional heads to capture the cohesions of the preference data, where ranking cohesion is defined as whether preference labels agree with derived rankings from predicted overall helpfulness of the responses, and overall helpfulness cohesion is defined as whether overall helpfulness labels agree with derived overall helpfulness from predicted rewards. Leveraging the cohesion signals from this reward model and through a neighborhood search, we proposed a novel prompt selection algorithm that selects prompts with the highest cohesion scores from their semantic neighborhood. This text-only prompt set, which includes multilingual sets, leads to significant gains in both auto benchmarks (4\% in Arena Hard~\cite{li2024crowdsourceddatahighqualitybenchmarks}, 7\% in win rate of AlpacaEval~\cite{dubois2025lengthcontrolledalpacaevalsimpleway} versus GPT4-Turbo, 10\% in Agent Sandbox~\cite{lu2025toolsandboxstatefulconversationalinteractive}, 7\% in GPQA~\cite{rein2023gpqagraduatelevelgoogleproofqa}, 5\% in Math500~\cite{lightman2023lets}) and human benchmarks (1.3-2.0\% increase in overall satisfaction across various locales) compared with the previous version of a text-only dataset for a text-only model.

\subsection{Tool-use \dri{Guoli Yin}}

Tool-use data collection poses a significant challenge due to its multi-turn nature and inherent software (tool) dependencies.
To facilitate efficient and high quality tool-use data collection, we designed a process-supervision annotation method as follows:
We created a tool-use agent annotation platform consisting of a reference model and a tool execution environment. The environment contains a curated selection of state-full and state-less tools and realistic human-authored databases as its initial states.
Annotators start by issuing a user query to the platform, which triggers a sequence of tool calls initiated by the agent model 
until the model decides to respond to the user.
The platform returns the entire trajectory, including the tool invocation details and the corresponding tool execution responses, as well as the final response for the user.
The platform supports reverting back to any turn in the dialog history, including resetting database states, allowing annotators to inspect model predictions, correct errors, and resume execution after the correction.
Annotators can provide follow-up requests until task completion.
In the end, the annotation process yields a tree-structured dataset, with a valid multi-turn tool-use SFT trajectory on the main stem and abandoned tool-use attempts as branches.

\subsection{Multilingual}

We extended language support to new locales and languages, adding these languages in phases to ensure high quality while avoiding regression in previously supported languages. 

By default, we induced the behavior that matches the output language to the input. However, we also enable the option to use different languages for prompts and responses by creating a small but diverse dataset with mixed languages that represents 0.4\% of the multilingual SFT mixture.

We introduced multilingual data in both the SFT and RLHF stages. In each stage, we sampled English and multilingual datasets in an 80:20 proportion. Data sets for both stages include a combination of human-written datasets and synthetic datasets. In human evaluations we find that RLHF provides significant lift over SFT, leading to 16:9 win/loss rate.

For hill-climbing on multilingual performance, we used the Instruction Following Eval (IFEval~\cite{ifeval}) and AlpacaEval with GPT4-o as a judge. We collected hundreds of prompts in each supported language written by native speakers because translated prompts were reported to be unnatural by our raters. With prompt tuning, we achieved good alignment between auto evals and human evals, enabling faster iteration.   

\section{Optimizations}
\label{sec:optimization}

We achieved an expansion of capabilities and quality improvements while \textit{increasing inference efficiency and reducing power consumption} of our on-device and Private Cloud Compute (PCC) server models~\cite{ApplePCC}. In addition to the architectural optimizations discussed above, we compressed the on-device model to 2 bits-per-weight using Quantization-Aware-Training (QAT), and we post-training compressed the server model to 3.56 bits-per-weight using Adaptive Scalable Texture Compression (ASTC). For both models, we quantized the embedding table to 4 bits per weight—using joint training with the base weights during QAT for the on-device model, and post-training quantization for the PCC model. We also quantized the KV-cache to 8 bits per weight and introduced low-rank adapters to recover quality lost during compression.

\paragraph{Quantization Aware Training (QAT)} is used to recover quality for the on-device model after compressing it to 2 bits-per-weight. During training, we simulate quantization effects by modifying the weight, $\mathbf{W}$, computation as:  
$$\tilde{\mathbf{W}} = s \cdot (\textit{clamp} (\lfloor \frac{\mathbf{W}}{s} + z\rceil, q_{min}, q_{max})- z),$$ 
where $s$ is the scaling factor, $z$ is the zero point, and $q_{min}, q_{max}$ define the quantization range. To handle the non-differentiable rounding operation during backpropagation, we use the straight-through estimator (STE) to approximate gradients.

Unlike the conventional quantization scheme which derives the scale from weights $\mathbf{W}$, we introduce a learnable scaling factor $f$ that adaptively fine-tunes the quantization range for each weight tensor. This factor scales the absolute maximum value of the tensor to compute the quantization scale as: $s = \frac{f \cdot \max(|\mathbf{W}|)}{q_{max}}$. Under aggressive 2-bit quantization, initializing $f$ is crucial for stable training. We propose an iterative method inspired by the Newton-Raphson algorithm to estimate a clipping scalar $c$ that better reflects the central weight distribution while mitigating the influence of outliers. In each iteration, weights above and below $c$ are rebalanced to iteratively refine the estimate. After convergence, the scaling factor is initialized as: $f_{init} = \frac{c}{\max(|\mathbf{W}|)}$. This initialization leads to more stable training dynamics and empirically accelerates convergence of the 2-bit model. 

QAT follows the same training stages as the base model but with fewer iterations. Unlike full-precision training, it is more sensitive to hyperparameter tuning. We found the AdamW optimizer~\citep{KingmaB14} to be more stable than Adafactor~\citep{ShazeerS18}, likely due to the need for accurate momentum estimation in low-bit regimes. Stability also improves with lower learning rates and gradient scaling inversely proportional to the square root of the neuron count. Additionally, we maintain an Exponential Moving Average (EMA) of the weights $\mathbf{W}$. This smoothing mechanism filters out noisy fluctuations in the weight trajectory, leading to more stable evaluation behavior and consistently improved metrics for the 2-bit model. Consistent with recent work~\citep{paretoQ}, we found a balanced 2-bit set \{-1.5, -0.5, 0.5, 1.5\} yields smoother training with fewer training loss spikes than an unbalanced set \{-2, -1, 0, 1\}. We set the weight decay to $0$ to encourage the model to utilize the full range of quantization levels, rather than collapsing weights around the zero point. 

\paragraph{Adaptive Scalable Texture Compression (ASTC)} is a block-based lossy compression format originally developed for efficient texture representation in GPU graphics pipelines~\cite{nystad2012adaptive}. The fixed-function hardware for ASTC decompression within Apple GPUs enables these blocks to be decoded on-the-fly with effectively zero compute cost from the perspective of the GPU shader or compute core. Leveraging this hardware for neural network inference enables efficient decompression of model weights without occupying general-purpose GPU compute resources.

We apply ASTC post-training to compress the final trained weights of the server model. Specifically, we use 6×6 blocks (i.e., 36 weights per block), encoded into 128-bit ASTC compressed values. This yields a storage footprint of approximately 3.56 bits-per-weight. We adopt the HDR-ch mode of ASTC, which supports higher precision encoding with non-negative 16-bit float channels. Since HDR-ch mode does not allow negative values, we perform a simple transformation: for each 6×6 block, we subtract the minimum value across all weights in that block, ensuring the data is strictly non-negative. This minimum value is stored separately as a single float-16 scalar alongside the compressed block. During inference, when the block is decompressed in hardware, the stored minimum is added back to the decompressed values before use. This transformation is lossless with respect to the min-shift and ensures correctness of downstream computations. The addition of the minimum value is fused into the corresponding tensor operations (e.g., matrix multiplications), ensuring that no separate reconstruction step is required and avoiding any runtime cost associated with the transformation.

This approach significantly reduces memory bandwidth and storage overhead which is particularly important for autoregressive decoding where weights are frequently accessed. Importantly, since decompression occurs transparently in hardware, there is no additional latency introduced during prompt-time inference.

\paragraph{Quality Recovery Adapters}
To recover model quality lost during the compression stages, we apply Low-Rank Adaptation (LoRA) adapters to both models and further fine-tune these adapters using the same data recipe as the base model training. These adapters introduce a small number of trainable parameters into each layer, allowing the model to adapt to the quantization and compression artifacts introduced in the base model compression. Because the core model weights remain frozen, the LoRA parameters effectively compensate for the lossy compression without requiring full model retraining. For the server model, before performing ASTC, we pull out the most significant singular vectors of the weights of the adapted layers into the LoRA adapter (e.g.~similar methods were presented by~\citet{meng2024pissa}). The residuals, which are the remaining singular vectors, make up the base model. Only the base model is then compressed using ASTC. This results in lower compression error, as ASTC now operates upon a smaller stack of singular vectors, and leaves the most significant singular vectors unchanged in the adapters.

\section{Foundation Models Framework}

    The new Foundation Models framework gives access to developers to start creating their own reliable, production-quality generative AI features with the approximately 3B parameter on-device language model. The $\sim$3B language foundation model at the core of Apple Intelligence excels at a diverse range of text tasks like summarization, entity extraction, text understanding, refinement, short dialog, generating creative content, and more. While we have specialized our on-device model for these tasks, it is not designed to be a chatbot for general world knowledge. We encourage app developers to use this framework to design helpful features tailored to their apps.

    To provide a simple and intuitive developer experience, the highlight of our framework is an idiomatic Swift-centric approach to constrained decoding called guided generation. Guided generation reduces the burden on developers for specifying a response format in their prompt, manually parsing strings, or dealing with the possibility of malformed model output. Instead, developers can directly generate rich Swift data structures by adding a  \verb|@Generable| macro annotation to Swift structs or enums. Guided generation uses a vertical integration with the model, the operating system, and developer tools. It begins with the Swift macros, which translate developer-defined data structures into a standardized response format specification. When prompting the model, the framework injects the response format into the prompt, and the model is able to understand and adhere to it because of post-training on a special dataset designed with the guided generation specification. Next, an OS daemon employs highly optimized, complimentary implementations of constrained decoding and speculative decoding to boost inference speed while ensuring that the model's output conforms to the expected format. Based on these outputs, the framework is able to create instances of Swift data structures from the model output. This streamlines the developer experience by letting app developers write much simpler code, backed by the Swift type system. The high level of abstraction of the framework provides flexibility for us to continue improving performance and generation quality under-the-hood. Backing guided generation and tool calling with a standardized schema representation means that we can continue to train our model for that format, and apps built on top of our framework will benefit from the improved performance.

    Tool calling offers developers the power to customize the $\sim$3B model's abilities by creating \textit{tools} that provide the model with specific kinds of information sources or services. The framework's approach to tool calling builds off of guided generation, where it can guarantee the structural correctness of tool calls by preventing hallucinated tool names or arguments. The developer provides an implementation of the simple \verb|Tool| Swift protocol, and the framework automatically and optimally handles the potentially complex call graphs of parallel and serial tool calls. Model post-training on tool-use data improved the model's reliability for this framework feature. 

    Next, the framework simplifies reasoning about performance and context by using an append-only, state-full session called \verb|LanguageModelSession|. Behind the scenes, this session is coupled to the model's key-value (KV) cache. Filling the KV cache can introduce latency, so the framework is built to prevent developers from unintentionally invalidating it. \verb|LanguageModelSession| supports streaming partially generated content as the model produces it leveraging the notion of snapshots. The framework handles the complex task of parsing partially generated output, and the developer simply sees streaming output as a Swift object that grows over time.

    We've carefully designed the framework to help app developers get the most out of the on-device model. For specialized use cases that require teaching the$\sim$3B model entirely new skills, we also provide a Python toolkit for training rank-32 LoRA adapters as well as optionally training a draft model for on-device speculative decoding. Adapters produced by the toolkit are fully compatible with the Foundation Models framework. However, each adapter is compatible with a single specific model version, meaning that a new adapter must be trained for each new version of the base model. Thus deploying one should be considered for advanced use cases after thoroughly exploring the capabilities of the base model. Since each adapter takes significant storage space, the Foundation Models framework leverages the Background Assets framework to download just a single adapter that matches the base model's version on device. To host adapters with Background Assets, developers have the flexibility to use their own server or use Apple's servers.

In addition to providing an easy-to-use, powerful Swift API, the Foundation Models framework comes with integrated tooling as part of the Xcode IDE. This includes a playground for prompt engineering using Swift code, a performance profiler for on-device model inference, and the ability to run the model directly in iOS and visionOS simulators. Read more about the framework in the developer documentation. The framework's approach to AI safety is discussed in the Responsible AI section below.
\section{Evaluation}

On pretraining benchmarks, we evaluated performance on Massive Multitask Language Understanding (MMLU), Multilingual Massive Multitask Language Understanding (MMMLU) and Multilingual Grade School Math (MGSM) and compared them to publicly accessible external models in \autoref{tab:on-device-metrics} and \autoref{tab:server-metrics}. We used the instruct and non-thinking version of the models and evaluated them only on the languages supported by AFM models with the Simple-Evals library~\citet{openai_simple_evals}. We found that AFM on-device model performs better than Qwen-2.5-3B, Gemma-3-4B and Gemma-3n-E4B on MMLU/MMMLU, but it lags slightly behind Gemma-3n-E4B on MGSM. AFM on-device model performs lower than the larger Qwen-3-4B model. AFM server models lag slightly to LLaMA 4 Scout, whose total size and active number of parameters are comparable, but has a bigger gap to larger models such as Qwen-3-235B and the proprietary GPT-4o.

\begin{table}[h]
\caption{AFM On-Device vs external models on representative benchmarks.}
\label{tab:on-device-metrics}
\centering
\vspace{0.5em}
\begin{tabular}{@{}lrrr@{}}
 \toprule
\textbf{Model} & \textbf{MMLU} & \textbf{MMMLU} & \textbf{MGSM} \\ \midrule
AFM On-Device & 67.85 & 60.60 & 74.91 \\
\qwen{2.5-3B} & 66.37 & 56.53 & 64.80 \\
\qwen{3-4B} & 75.10 & 66.52 & 82.97 \\
\gemma{3-4B} & 62.81 & 56.71 & 74.74 \\
\gemma{3n-E4B} & 57.84 & 50.93 & 77.77 \\
 \bottomrule
\end{tabular}

\end{table}

\begin{table}[h]
\caption{AFM Server vs external models on representative benchmarks.}
\label{tab:server-metrics}
\centering
\vspace{0.5em}
\begin{tabular}{@{}lrrr@{}}
 \toprule
\textbf{Model} & \textbf{MMLU} & \textbf{MMMLU} & \textbf{MGSM} \\ \midrule
AFM Server & 80.20 & 74.60 & 87.09 \\
\llama{4 Scout} & 84.88 & 80.24 & 90.34 \\
\qwen{3-235B} & 87.52 & 82.95 & 92.00 \\
\gpt{4o} & 85.70 & 84.00 & 90.30 \\
 \bottomrule
\end{tabular}

\end{table}

Our model optimizations deliver significantly higher token throughput, substantially lower latency, and remarkable reduced DRAM footprint during inference time, comparing to models using 16-bit weights. We carefully tailor these optimizations to best preserve the model quality. \autoref{tab:optimization_metrics} presents representative quality evaluations before and after applying all optimization techniques described in \autoref{sec:optimization}, including Quantization-Aware Training (QAT), ASTC, and quality recovery adapters. 

\begin{table}[h]
\caption{Representative metrics on quality impact through optimization.}
\vspace{0.5em}
\label{tab:optimization_metrics}
\centering
\resizebox{\textwidth}{!}{%

\begin{tabular}{lrr|r}
 \toprule
\textbf{Model} & \textbf{MMLU} & \textbf{IFEval (instruct)} & \textbf{Bits-per-weight} \\ \midrule
AFM On-Device & 67.8 & 85.1 & 16 \\
AFM On-Device Opt & 64.4 & 82.3 & 2 \\
\midrule
AFM Server & 80.0 & 89.1 & 16 \\
AFM Server Opt & 79.2 & 90.2 & 3.6 \\
 \bottomrule
\end{tabular}
}
\vspace{10pt}
\end{table}

Further, we conducted quality evaluations of our on-device and server-based models offline using human graders. We evaluate along standard fundamental language and reasoning capabilities, including Analytical Reasoning, Brainstorming, Chat, Classification, Closed Question and Answering, Coding, Creative Writing, Extraction, Mathematical Reasoning, Open Question and Answering, Rewriting, Summarization, and Tool-use.

As we expanded our model support to additional languages and locales, we increased our evaluation task set to be locale specific. Human graders assessed the model's ability to produce a response that was native-sounding to a user in that locale. For example, a model responding to an English sports question from a user in Great Britain is expected to know "football" is a more locally appropriate term than "soccer". Graders could flag the model's response for a number of issues, including un-localized terms or unnatural phrases. Locale specific evaluations used similar categories as US English, with the exception of locale-agnostic domains like math and coding.

\begin{figure}[!t]
    \includegraphics[width=\linewidth,alt={Chart showing which models win, tie, or loose in a direct comparison for text prompts. On-device comparisons are on the left and server comparisons are on the bottom.},center]{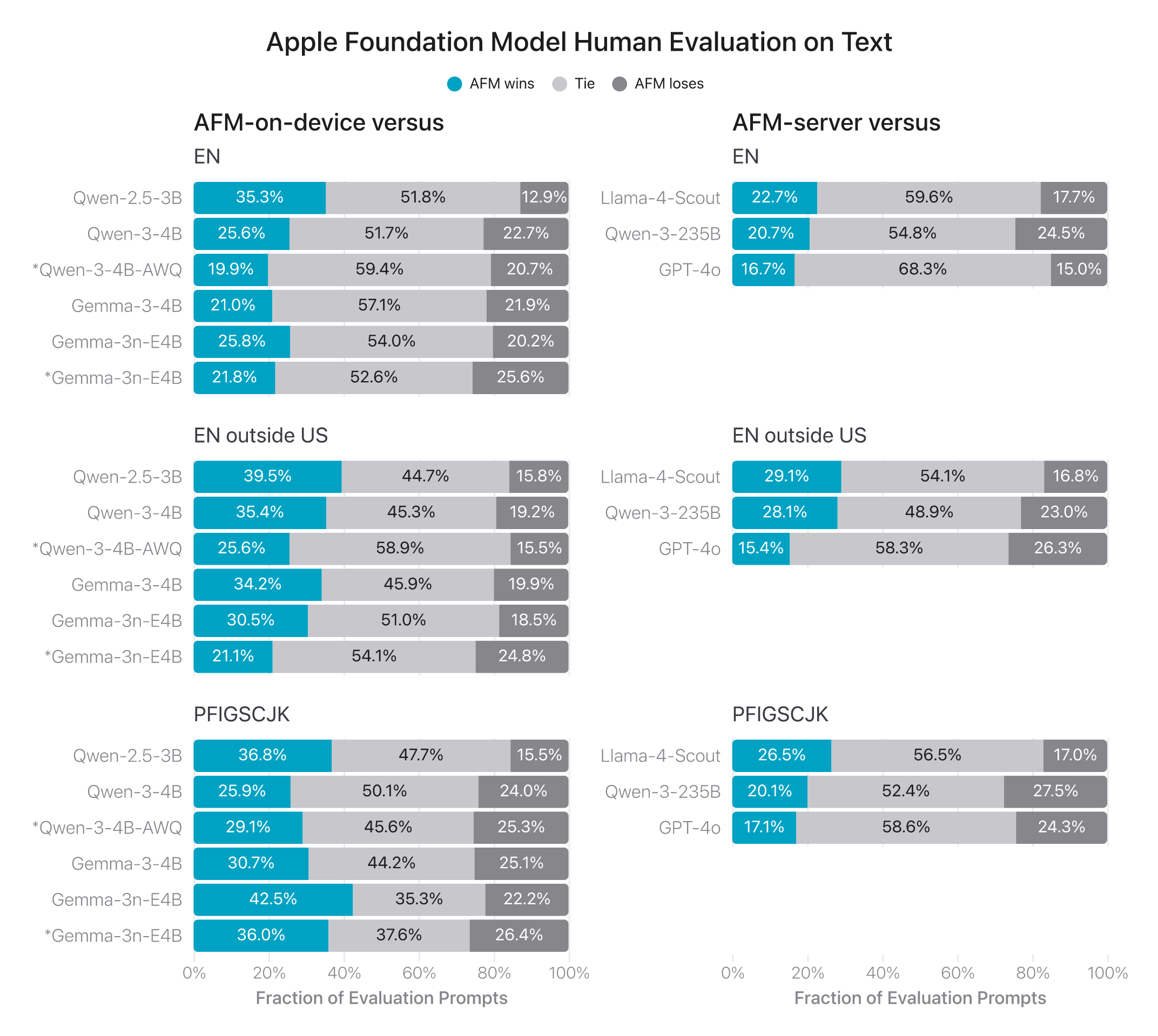}
    \vspace{-1.5em}
    \caption{Fraction of preferred responses in side-by-side evaluation of text responses comparing Apple's foundation model against publicly accessible models.
    Results are presented across 3 locale groups, a lens by which we view Apple Intelligence's internationalization.
    English outside of the US for example includes English in Great Britain and English in Canada, among others.
    PFIGSCJK refers to the languages Portuguese, French, Italian, Spanish, Chinese (Simplified), Japanese, and Korean. 
    \textit{\textsuperscript{*}Denotes models tested against Apple on-device compressed model.}
    }
    \label{fig:text-eval}
\end{figure}

We compared our models to publicly accessible external models, namely
\qwen{2.5-3B}, \qwen{3-4B}, \gemma{3-4B}, \gemma{3n},
\llama{4 Scout}, \qwen{3-235B}, and OpenAI's \gpt{4o}. 
We found that our on-device model performs favorably against the slightly larger \qwen{2.5-3B} and \gemma{3n} and is competitive against the larger \qwen{3-4B} and \gemma{3-4B},
and our server-based model performs favorably against \llama{4 Scout}, whose total size and active number of parameters are comparable to our server model, but is behind larger models such as \qwen{3-235B} and the proprietary \gpt{4o}.

With our model support expanding to the image modality, an evaluation set of Image-Question pairs was used to assess Image Understanding capabilities. This evaluation set contained similar categories as the text evaluation set, along with image specific categories like Infographics, which challenge the model to reason about text rich images. We compared the on-device model to vision models of similar size, namely InternVL-2.5-4B, \qwen{2.5-VL-3B}, and \gemma{3-4B}, and our server model to \llama{4 Scout}, \qwen{2.5-VL-32B}, and \gpt{-4o}. We found that Apple's on-device model performs favorably against the larger InternVL and Qwen and competitively against Gemma, and our server model outperforms \qwen{2.5-VL}, at less than half of inference FLOPS, but is behind the larger \llama{4 Scout} and \gpt{-4o}.

\begin{figure}[!ht]
    \includegraphics[width=\linewidth,alt={Chart showing which models win, tie, or loose in a direct comparison for image prompts. On-device comparisons are on the left and server comparisons are on the bottom.},center]{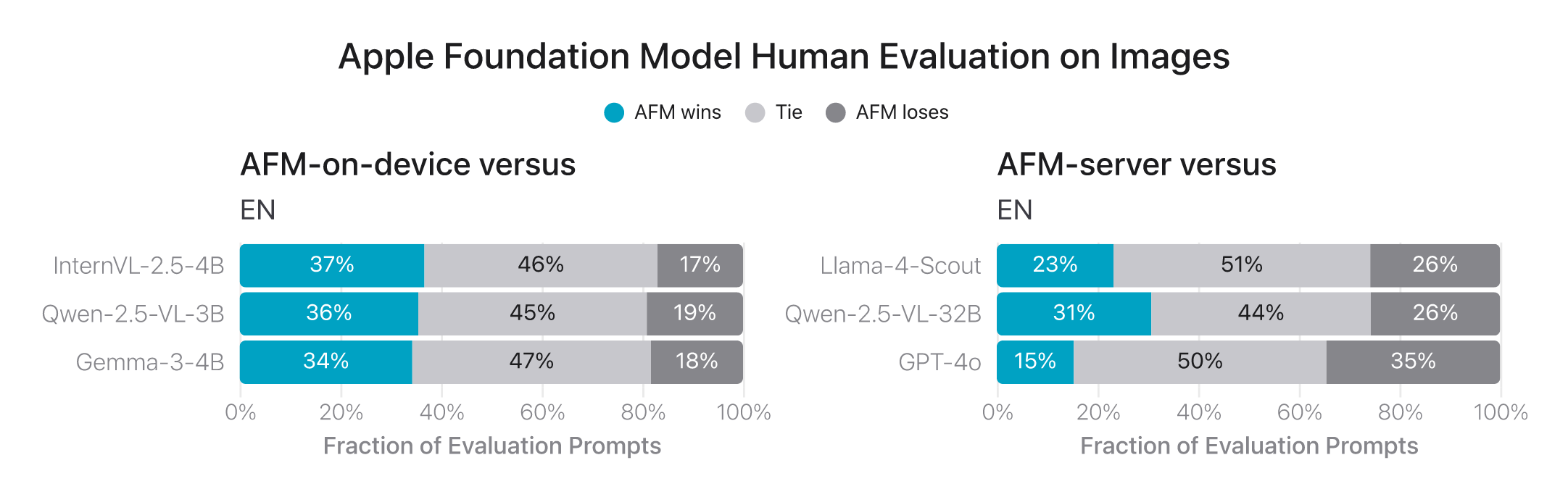}
    \vspace{-1.5em}
    \caption{Fraction of preferred responses in side-by-side evaluation of image responses comparing Apple's foundation model against comparable models.
    }
    \label{fig:image-eval}
\end{figure}

Since these models are developed to deliver a wide range of features that can help our users across everyday tasks, we perform detailed evaluation for each specific feature, on top of base model evaluations. For example, consider the Visual Intelligence feature which creates a calendar event from an image of a flyer. An evaluation set of flyers is collected across a broad range of environmental settings, camera angles, and other difficulties. This is used to assess the model's ability to accurately extract information from the flyer, including the date and location, to properly create the calendar event.

\section{Responsible AI}

Apple Intelligence is designed with our core values at every step and built on a foundation of groundbreaking privacy innovations. At every stage of developing and advancing Apple Intelligence, we use our Responsible AI principles to guide our features and models:

\begin{enumerate}
    \item \textbf{Empower users with intelligent tools:} We identify areas where AI can be used responsibly to create tools for addressing specific user needs. We respect how our users choose to use these tools to accomplish their goals.
    \item \textbf{Represent our users:} We build deeply personal products with the goal of representing users around the globe authentically. We work continuously to avoid perpetuating stereotypes and systemic biases across our AI tools and models.
    \item \textbf{Design with care:} We take precautions at every stage of our process, including design, model training, feature development, and quality evaluation to identify how our AI tools may be misused or lead to potential harm. We will continuously monitor and proactively improve our AI tools with the help of user feedback.
    \item \textbf{Protect privacy:} We protect our users' privacy with powerful on-device processing and groundbreaking infrastructure like Private Cloud Compute. We do not use our users' private personal data or user interactions when training our foundation models.
\end{enumerate}

These principles guide our work throughout the product development cycle, informing our product design, policies, evaluations, and mitigations. As part of Apple's commitment to responsible AI, we have continued to identify and mitigate the risks inherent to the use of foundation models, such as hallucinations and susceptibility to prompt injections. Our safety taxonomy helps us identify sensitive content that should be handled with care. The taxonomy is updated regularly based on ongoing risk assessments, model capabilities, and internal and external human evaluations. It currently contains 6 categories and 58 subcategories, which includes areas such as Slurs and Hate Speech/Symbols, Controversial and Sensitive Topics, and Algorithmic Biases and Stereotypes.

To evaluate the safety of Apple Intelligence, we assessed both the foundation models as well as each feature that uses the models prior to deployment. For foundation models, we combined internal and external human evaluation with auto-grading, and compared our models to external models for benchmarking. We constructed targeted safety evaluation datasets to assess the performance of the foundation model on tasks such as summarization, question-answering, and brainstorming, as it applies to high-risk and sensitive content. For individual features, we designed datasets that focus on user-facing risks to specifically identify unwanted or unintended outcomes. We also look at how potential gaps in quality can exacerbate risks when applied to sensitive app-specific content. For example, we took care in designing the new Foundation Models framework and supporting resources to help improve generative AI safety for apps. The framework enforces a base level of safety with built-in safety guardrails to mitigate harmful model input and output. To help app designers and developers incorporate AI safety that is tailored to their Apps, we created educational resources, such as new \href{https://developer.apple.com/design/human-interface-guidelines/generative-ai}{Generative AI Human Interface Guidelines} for Responsible AI principles.

As we expand our features to new languages, we expand our safety representation across regions and cultures, and we continue to make improvements to account for the wide cultural and linguistic diversity of our users. In addition to adhering to local laws and regulations, we leverage a combination of high-quality external representative data sources, engaged with internal and external legal, language, and cultural experts, as well as reviewed precedents from previous product decisions to ensure that our approach was contextually respectful and relevant. To design our mitigation steps for multilingual use, we began with multilingual post-training alignment at the foundational model level, then extended to feature-specific adapters that integrate safety alignment data. Additionally, we expanded our guardrail models, designed to intercept harmful prompts, with language-specific training data while maintaining a multilingual adapter. We developed customized datasets to mitigate culture-specific risks and biases and stereotypes in model outputs. Similarly, we extended our evaluation datasets across languages and locales with tools such as machine translation and targeted synthetic data generation, all refined by native speakers. Finally, we conducted human red teaming across features to identify risks unique to each locale. While we work to mitigate newly discovered risks in subsequent versions of our foundational models, we may add related terms to an overrides list for special handling to prevent harmful outputs; for example, a slur in a prompt may trigger a warning in writing tools or prevent the generation of a Genmoji.

We continuously monitor and proactively improve our features with the help of submitted user feedback. Feedback from users is a critical component of our Responsible AI approach. In Image Playground, for instance, users can provide feedback on generated images by tapping "thumbs up" or "thumbs down", with the option to add comments. App developers can similarly offer feedback through \href{https://developer.apple.com/bug-reporting/}{Feedback Assistant}. Feedback from users and developers, along with evaluation data and other metrics, helps us continuously improve Apple Intelligence features and models.
\section{Conclusion}

We are excited to make the language foundation models at the core of Apple Intelligence more efficient and more capable, unlocking a wide range of helpful features integrated across our software platforms, and available to our users around the globe across many languages. We are also giving app developers direct access to our on-device language foundation model with a new Foundation Models framework. App developers can take advantage of AI inference that is free of cost and accessible with just a few lines of code, and bring capabilities such as text extraction and summarization to their apps. The latest advancements in Apple Intelligence models continue to draw on our core values, like our commitment to privacy, as well as our Responsible AI approach.

\bibliography{main}

\begin{thebibliography}{31}
\providecommand{\natexlab}[1]{#1}
\providecommand{\url}[1]{\texttt{#1}}
\expandafter\ifx\csname urlstyle\endcsname\relax
  \providecommand{\doi}[1]{doi: #1}\else
  \providecommand{\doi}{doi: \begingroup \urlstyle{rm}\Url}\fi

\bibitem[Ahmadian et~al.(2024)Ahmadian, Cremer, Gall{\'e}, Fadaee, Kreutzer, {\"U}st{\"u}n, and Hooker]{ahmadian2024back}
Arash Ahmadian, Chris Cremer, Matthias Gall{\'e}, Marzieh Fadaee, Julia Kreutzer, Ahmet {\"U}st{\"u}n, and Sara Hooker.
\newblock Back to basics: Revisiting {REINFORCE} style optimization for learning from human feedback in {LLMs}.
\newblock 2024.

\bibitem[Apple(2024)]{ApplePCC}
Apple.
\newblock Private cloud compute: A new frontier for {AI} privacy in the cloud.
\newblock \url{https://security.apple.com/blog/private-cloud-compute/}, 2024.
\newblock Accessed: 2025-07-01.

\bibitem[Beltagy et~al.(2020)Beltagy, Peters, and Cohan]{Beltagy2020Longformer}
Iz~Beltagy, Matthew~E. Peters, and Arman Cohan.
\newblock Longformer: The long-document transformer, 2020.
\newblock URL \url{https://arxiv.org/abs/2004.05150}.

\bibitem[Dosovitskiy et~al.(2020)Dosovitskiy, Beyer, Kolesnikov, Weissenborn, Zhai, Unterthiner, Dehghani, Minderer, Heigold, Gelly, et~al.]{dosovitskiy2020image}
Alexey Dosovitskiy, Lucas Beyer, Alexander Kolesnikov, Dirk Weissenborn, Xiaohua Zhai, Thomas Unterthiner, Mostafa Dehghani, Matthias Minderer, Georg Heigold, Sylvain Gelly, et~al.
\newblock An image is worth 16x16 words: Transformers for image recognition at scale.
\newblock \emph{arXiv preprint arXiv:2010.11929}, 2020.

\bibitem[Du et~al.(2022)Du, Huang, Dai, Tong, Lepikhin, Xu, Krikun, Zhou, Yu, Firat, Zoph, Fedus, Bosma, Zhou, Wang, Wang, Webster, Pellat, Robinson, Meier-Hellstern, Duke, Dixon, Zhang, Le, Wu, Chen, and Cui]{du2022}
Nan Du, Yanping Huang, Andrew~M. Dai, Simon Tong, Dmitry Lepikhin, Yuanzhong Xu, Maxim Krikun, Yanqi Zhou, Adams~Wei Yu, Orhan Firat, Barret Zoph, Liam Fedus, Maarten Bosma, Zongwei Zhou, Tao Wang, Yu~Emma Wang, Kellie Webster, Marie Pellat, Kevin Robinson, Kathleen Meier-Hellstern, Toju Duke, Lucas Dixon, Kun Zhang, Quoc~V Le, Yonghui Wu, Zhifeng Chen, and Claire Cui.
\newblock {GLaM}: Efficient scaling of language models with mixture-of-experts.
\newblock 2022.
\newblock URL \url{https://arxiv.org/abs/2112.06905}.

\bibitem[Du et~al.(2024)Du, Gunter, Kong, Lee, Wang, Zhang, Du, and Pang]{axlearnmoe}
Xianzhi Du, Tom Gunter, Xiang Kong, Mark Lee, Zirui Wang, Aonan Zhang, Nan Du, and Ruoming Pang.
\newblock Revisiting moe and dense speed-accuracy comparisons for llm training, 2024.
\newblock URL \url{https://arxiv.org/pdf/2405.15052}.

\bibitem[Dubois et~al.(2025)Dubois, Galambosi, Liang, and Hashimoto]{dubois2025lengthcontrolledalpacaevalsimpleway}
Yann Dubois, Balázs Galambosi, Percy Liang, and Tatsunori~B. Hashimoto.
\newblock Length-controlled alpacaeval: A simple way to debias automatic evaluators, 2025.
\newblock URL \url{https://arxiv.org/abs/2404.04475}.

\bibitem[Gale et~al.(2022)Gale, Narayanan, Young, and Zaharia]{megablox}
Trevor Gale, Deepak Narayanan, Cliff Young, and Matei Zaharia.
\newblock {MegaBlocks}: Efficient sparse training with mixture-of-experts.
\newblock 2022.
\newblock URL \url{https://arxiv.org/abs/2211.15841}.

\bibitem[Gunter et~al.(2024)Gunter, Wang, Wang, Pang, Narayanan, Zhang, Zhang, Chen, Chiu, Qiu, et~al.]{gunter2024apple}
Tom Gunter, Zirui Wang, Chong Wang, Ruoming Pang, Andy Narayanan, Aonan Zhang, Bowen Zhang, Chen Chen, Chung-Cheng Chiu, David Qiu, et~al.
\newblock {Apple Intelligence} foundation language models.
\newblock \emph{arXiv preprint arXiv:2407.21075}, 2024.

\bibitem[Kingma and Ba(2015)]{KingmaB14}
Diederik~P. Kingma and Jimmy Ba.
\newblock Adam: {A} method for stochastic optimization.
\newblock In Yoshua Bengio and Yann LeCun, editors, \emph{3rd International Conference on Learning Representations, {ICLR} 2015, San Diego, CA, USA, May 7-9, 2015, Conference Track Proceedings}, 2015.
\newblock URL \url{http://arxiv.org/abs/1412.6980}.

\bibitem[Komatsuzaki et~al.(2023)Komatsuzaki, Puigcerver, Lee-Thorp, Ruiz, Mustafa, Ainslie, Tay, Dehghani, and Houlsby]{komatsuzaki2023sparse}
Aran Komatsuzaki, Joan Puigcerver, James Lee-Thorp, Carlos~Riquelme Ruiz, Basil Mustafa, Joshua Ainslie, Yi~Tay, Mostafa Dehghani, and Neil Houlsby.
\newblock Sparse upcycling: Training mixture-of-experts from dense checkpoints.
\newblock In \emph{The Eleventh International Conference on Learning Representations}, 2023.
\newblock URL \url{https://openreview.net/forum?id=T5nUQDrM4u}.

\bibitem[Lee et~al.(2025)Lee, Gunter, Lan, Peebles, Zhou, Zou, Bangalore, Chiu, Du, Du, Dufter, Hou, Huang, Hwang, Kong, Lei, Lei, Li, Li, Lu, Lu, Ma, Qiu, Rathod, Tong, Tu, Wang, Wang, Wang, Weers, Wiseman, Yin, Zhang, Zhou, Zhuo, Leong, and Pang]{axlearn}
Mark Lee, Tom Gunter, Chang Lan, John Peebles, Hanzhi Zhou, Kelvin Zou, Sneha Bangalore, Chung-Cheng Chiu, Nan Du, Xianzhi Du, Philipp Dufter, Ruixuan Hou, Haoshuo Huang, Dongseong Hwang, Xiang Kong, Jinhao Lei, Tao Lei, Meng Li, Li~Li, Jiarui Lu, Zhiyun Lu, Yiping Ma, David Qiu, Vivek Rathod, Senyu Tong, Zhucheng Tu, Jianyu Wang, Yongqiang Wang, Zirui Wang, Floris Weers, Sam Wiseman, Guoli Yin, Bowen Zhang, Xiyou Zhou, Danyang Zhuo, Cheng Leong, and Ruoming Pang.
\newblock Axlearn: Modular large model training on heterogeneous infrastructure, 2025.
\newblock URL \url{https://arxiv.org/abs/2507.05411}.

\bibitem[Lepikhin et~al.(2021)Lepikhin, Lee, Xu, Chen, Firat, Huang, Krikun, Shazeer, and Chen]{gshard}
Dmitry Lepikhin, HyoukJoong Lee, Yuanzhong Xu, Dehao Chen, Orhan Firat, Yanping Huang, Maxim Krikun, Noam Shazeer, and Zhifeng Chen.
\newblock {GShard}: Scaling giant models with conditional computation and automatic sharding.
\newblock 2021.
\newblock URL \url{https://openreview.net/forum?id=qrwe7XHTmYb}.

\bibitem[Li et~al.(2024)Li, Chiang, Frick, Dunlap, Wu, Zhu, Gonzalez, and Stoica]{li2024crowdsourceddatahighqualitybenchmarks}
Tianle Li, Wei-Lin Chiang, Evan Frick, Lisa Dunlap, Tianhao Wu, Banghua Zhu, Joseph~E. Gonzalez, and Ion Stoica.
\newblock From crowdsourced data to high-quality benchmarks: Arena-hard and benchbuilder pipeline, 2024.
\newblock URL \url{https://arxiv.org/abs/2406.11939}.

\bibitem[Li et~al.(2022)Li, Mao, Girshick, and He]{li2022exploring}
Yanghao Li, Hanzi Mao, Ross Girshick, and Kaiming He.
\newblock Exploring plain vision transformer backbones for object detection.
\newblock In \emph{European conference on computer vision}, pages 280--296. Springer, 2022.

\bibitem[Li et~al.(2023)Li, Fan, Hu, Feichtenhofer, and He]{li2023scaling}
Yanghao Li, Haoqi Fan, Ronghang Hu, Christoph Feichtenhofer, and Kaiming He.
\newblock Scaling language-image pre-training via masking.
\newblock In \emph{Proceedings of the IEEE/CVF conference on computer vision and pattern recognition}, pages 23390--23400, 2023.

\bibitem[Lightman et~al.(2023)Lightman, Kosaraju, Burda, Edwards, Baker, Lee, Leike, Schulman, Sutskever, and Cobbe]{lightman2023lets}
Hunter Lightman, Vineet Kosaraju, Yura Burda, Harri Edwards, Bowen Baker, Teddy Lee, Jan Leike, John Schulman, Ilya Sutskever, and Karl Cobbe.
\newblock Let's verify step by step.
\newblock \emph{arXiv preprint arXiv:2305.20050}, 2023.

\bibitem[Liu et~al.(2025)Liu, Zhao, Huang, Chen, Zhang, Zhao, Roy, Jin, Xiong, Shi, Xiao, Tian, Soran, Krishnamoorthi, Blankevoort, and Chandra]{paretoQ}
Zechun Liu, Changsheng Zhao, Hanxian Huang, Sijia Chen, Jing Zhang, Jiawei Zhao, Scott Roy, Lisa Jin, Yunyang Xiong, Yangyang Shi, Lin Xiao, Yuandong Tian, Bilge Soran, Raghuraman Krishnamoorthi, Tijmen Blankevoort, and Vikas Chandra.
\newblock {ParetoQ}: Scaling laws in extremely low-bit {LLM} quantization.
\newblock \emph{CoRR}, abs/2502.02631, 2025.
\newblock \doi{10.48550/ARXIV.2502.02631}.
\newblock URL \url{https://doi.org/10.48550/arXiv.2502.02631}.

\bibitem[Lu et~al.(2025)Lu, Holleis, Zhang, Aumayer, Nan, Bai, Ma, Ma, Li, Yin, Wang, and Pang]{lu2025toolsandboxstatefulconversationalinteractive}
Jiarui Lu, Thomas Holleis, Yizhe Zhang, Bernhard Aumayer, Feng Nan, Felix Bai, Shuang Ma, Shen Ma, Mengyu Li, Guoli Yin, Zirui Wang, and Ruoming Pang.
\newblock Toolsandbox: A stateful, conversational, interactive evaluation benchmark for llm tool use capabilities, 2025.
\newblock URL \url{https://arxiv.org/abs/2408.04682}.

\bibitem[Meng et~al.(2024)Meng, Wang, and Zhang]{meng2024pissa}
Fanxu Meng, Zhaohui Wang, and Muhan Zhang.
\newblock Pissa: Principal singular values and singular vectors adaptation of large language models.
\newblock \emph{Advances in Neural Information Processing Systems}, 37:\penalty0 121038--121072, 2024.

\bibitem[Nystad et~al.(2012)Nystad, Lassen, Pomianowski, Ellis, and Olson]{nystad2012adaptive}
J{\"o}rn Nystad, Anders Lassen, Andy Pomianowski, Sean Ellis, and Tom Olson.
\newblock Adaptive scalable texture compression.
\newblock In \emph{Proceedings of the Fourth ACM SIGGRAPH/Eurographics Conference on High-Performance Graphics}, pages 105--114, 2012.

\bibitem[{OpenAI}(2025)]{openai_simple_evals}
{OpenAI}.
\newblock {simple‑evals}: A lightweight library for evaluating language models.
\newblock \url{https://github.com/openai/simple-evals}, 2025.
\newblock Accessed: 2025-07-13.

\bibitem[Radford et~al.(2021)Radford, Kim, Hallacy, Ramesh, Goh, Agarwal, Sastry, Askell, Mishkin, Clark, et~al.]{radford2021learning}
Alec Radford, Jong~Wook Kim, Chris Hallacy, Aditya Ramesh, Gabriel Goh, Sandhini Agarwal, Girish Sastry, Amanda Askell, Pamela Mishkin, Jack Clark, et~al.
\newblock Learning transferable visual models from natural language supervision.
\newblock In \emph{International conference on machine learning}, pages 8748--8763. PmLR, 2021.

\bibitem[Rein et~al.(2023)Rein, Hou, Stickland, Petty, Pang, Dirani, Michael, and Bowman]{rein2023gpqagraduatelevelgoogleproofqa}
David Rein, Betty~Li Hou, Asa~Cooper Stickland, Jackson Petty, Richard~Yuanzhe Pang, Julien Dirani, Julian Michael, and Samuel~R. Bowman.
\newblock Gpqa: A graduate-level google-proof q\&a benchmark, 2023.
\newblock URL \url{https://arxiv.org/abs/2311.12022}.

\bibitem[Shazeer and Stern(2018)]{ShazeerS18}
Noam Shazeer and Mitchell Stern.
\newblock Adafactor: Adaptive learning rates with sublinear memory cost.
\newblock In Jennifer~G. Dy and Andreas Krause, editors, \emph{Proceedings of the 35th International Conference on Machine Learning, {ICML} 2018, Stockholmsm{\"{a}}ssan, Stockholm, Sweden, July 10-15, 2018}, volume~80 of \emph{Proceedings of Machine Learning Research}, pages 4603--4611. {PMLR}, 2018.
\newblock URL \url{http://proceedings.mlr.press/v80/shazeer18a.html}.

\bibitem[Shazeer et~al.(2017)Shazeer, Mirhoseini, Maziarz, Davis, Le, Hinton, and Dean]{noam2017}
Noam Shazeer, Azalia Mirhoseini, Krzysztof Maziarz, Andy Davis, Quoc Le, Geoffrey Hinton, and Jeff Dean.
\newblock Outrageously large neural networks: The sparsely-gated mixture-of-experts layer.
\newblock 2017.
\newblock URL \url{https://arxiv.org/abs/1701.06538}.

\bibitem[Snowflake(2025)]{SwiftKV}
Snowflake.
\newblock Swiftkv: Accelerating enterprise llm workloads with knowledge preserving compute reduction.
\newblock \url{https://www.snowflake.com/en/engineering-blog/swiftkv-llm-compute-reduction/}, 2025.
\newblock Accessed: 2025-07-16.

\bibitem[Sun et~al.(2024)Sun, Dong, Zhu, Huang, Wang, Ma, Zhang, Wang, and Wei]{sun2024you}
Yutao Sun, Li~Dong, Yi~Zhu, Shaohan Huang, Wenhui Wang, Shuming Ma, Quanlu Zhang, Jianyong Wang, and Furu Wei.
\newblock You only cache once: Decoder-decoder architectures for language models.
\newblock \emph{Advances in Neural Information Processing Systems}, 37:\penalty0 7339--7361, 2024.

\bibitem[Yang et~al.(2025)Yang, Venkitesh, Talupuru, Lin, Cairuz, Blunsom, and Locatelli]{yang2025ropenopeagainnew}
Bowen Yang, Bharat Venkitesh, Dwarak Talupuru, Hangyu Lin, David Cairuz, Phil Blunsom, and Acyr Locatelli.
\newblock Rope to nope and back again: A new hybrid attention strategy, 2025.
\newblock URL \url{https://arxiv.org/abs/2501.18795}.

\bibitem[Zhou et~al.(2023)Zhou, Lu, Mishra, Brahma, Basu, Luan, Zhou, and Hou]{ifeval}
Jeffrey Zhou, Tianjian Lu, Swaroop Mishra, Siddhartha Brahma, Sujoy Basu, Yi~Luan, Denny Zhou, and Le~Hou.
\newblock Instruction-following evaluation for large language models.
\newblock 2023.

\bibitem[Zoph et~al.(2022)Zoph, Bello, Kumar, Du, Huang, Dean, Shazeer, and Fedus]{zoph2022}
Barret Zoph, Irwan Bello, Sameer Kumar, Nan Du, Yanping Huang, Jeff Dean, Noam Shazeer, and William Fedus.
\newblock {ST-MoE}: Designing stable and transferable sparse expert models.
\newblock 2022.
\newblock URL \url{https://arxiv.org/abs/2202.08906}.

\end{thebibliography}
\newpage
\appendix
\section*{Contributors}
\label{sec:contributors}

We thank the following contributors for their work on this project: (in random order)

\begin{multicols}{2}
\raggedright
Ethan Li\\
Anders Boesen Lindbo Larsen\\
Chen Zhang\\
Xiyou Zhou\\
Jun Qin\\
Dian Ang Yap\\
Narendran Raghavan\\
Xuankai Chang\\
Margit Bowler\\
Eray Yildiz\\
John Peebles\\
Hannah Gillis Coleman\\
Matteo Ronchi\\
Peter Gray\\
Keen You\\
Anthony Spalvieri-Kruse\\
Ruoming Pang\\
Reed Li\\
Yuli Yang\\
Emad Soroush\\
Zhiyun Lu\\
Crystal Xiao\\
Rong Situ\\
Jordan Huffaker\\
David Griffiths\\
Zaid Ahmed\\
Peng Zhang\\
Daniel Parilla\\
Asaf Liberman\\
Jennifer Mallalieu\\
Parsa Mazaheri\\
Qibin Chen\\
Manjot Bilkhu\\
Aonan Zhang\\
Eric Wang\\
Dave Nelson\\
Michael FitzMaurice\\
Thomas Voice\\
Jeremy Liu\\
Josh Shaffer\\
Shiwen Zhao\\
Prasanth Yadla\\
Farzin Rasteh\\
Pengsheng Guo\\
Arsalan Farooq\\
Jeremy Snow\\
Stephen Murphy\\
Tao Lei\\
Minsik Cho\\
George Horrell\\
Sam Dodge\\
Lindsay Hislop\\
Sumeet Singh\\
Alex Dombrowski\\
Aiswarya Raghavan\\
Sasha Sirovica\\
Mandana Saebi\\
Faye Lao\\
Max Lam\\
TJ Lu\\
Zhaoyang Xu\\
Karanjeet Singh\\
Marc Kirchner\\
David Mizrahi\\
Rajat Arora\\
Haotian Zhang\\
Henry Mason\\
Lawrence Zhou\\
Yi Hua\\
Ankur Jain\\
Felix Bai\\
Joseph Astrauskas\\
Floris Weers\\
Josh Gardner\\
Mira Chiang\\
Yi Zhang\\
Pulkit Agrawal\\
Tony Sun\\
Quentin Keunebroek\\
Matthew Hopkins\\
Bugu Wu\\
Tao Jia\\
Chen Chen\\
Xingyu Zhou\\
Nanzhu Wang\\
Peng Liu\\
Ruixuan Hou\\
Rene Rauch\\
Yuan Gao\\
Afshin Dehghan\\
Jonathan Janke\\
Zirui Wang\\
Cha Chen\\
Xiaoyi Ren\\
Feng Nan\\
Josh Elman\\
Dong Yin\\
Yusuf Goren\\
Jeff Lai\\
Yiran Fei\\
Syd Evans\\
Muyang Yu\\
Guoli Yin\\
Yi Qin\\
Erin Feldman\\
Isha Garg\\
Aparna Rajamani\\
Karla Vega\\
Walker Cheng\\
TJ Collins\\
Hans Han\\
Raul Rea Menacho\\
Simon Yeung\\
Sophy Lee\\
Phani Mutyala\\
Ying-Chang Cheng\\
Zhe Gan\\
Sprite Chu\\
Justin Lazarow\\
Alessandro Pappalardo\\
Federico Scozzafava\\
Jing Lu\\
Erik Daxberger\\
Laurent Duchesne\\
Jen Liu\\
David Güera\\
Stefano Ligas\\
Mary Beth Kery\\
Brent Ramerth\\
Ciro Sannino\\
Marcin Eichner\\
Haoshuo Huang\\
Rui Qian\\
Moritz Schwarzer-Becker\\
David Riazati\\
Mingfei Gao\\
Bailin Wang\\
Jack Cackler\\
Yang Lu\\
Ransen Niu\\
John Dennison\\
Guillaume Klein\\
Jeffrey Bigham\\
Deepak Gopinath\\
Navid Shiee\\
Darren Botten\\
Guillaume Tartavel\\
Alex Guillen Garcia\\
Sam Xu\\
Victoria MönchJuan Haladjian\\
Zi-Yi Dou\\
Matthias Paulik\\
Adolfo Lopez Mendez\\
Zhen Li\\
Hong-You Chen\\
Chao Jia\\
Dhaval Doshi\\
Zhengdong Zhang\\
Raunak Manjani\\
Aaron Franklin\\
Zhile Ren\\
David Chen\\
Artsiom Peshko\\
Nandhitha Raghuram\\
Hans Hao\\
Jiulong Shan\\
Kavya Nerella\\
Ramsey Tantawi\\
Vivek Kumar\\
Saiwen Wang\\
Brycen Wershing\\
Bhuwan Dhingra\\
Dhruti Shah\\
Ob Adaranijo\\
Xin Zheng\\
Tait Madsen\\
Hadas Kotek\\
Chang Liu\\
Yin Xia\\
Hanli Li\\
Suma Jayaram\\
Yanchao Sun\\
Ahmed Fakhry\\
Vasileios Saveris\\
Dustin Withers\\
Yanghao Li\\
Alp Aygar\\
Andres Romero Mier Y Teran\\
Kaiwei Huang\\
Mark Lee\\
Xiujun Li\\
Yuhong Li\\
Tyler Johnson\\
Jay Tang\\
Joseph Yitan Cheng\\
Futang Peng\\
Andrew Walkingshaw\\
Lucas Guibert\\
Abhishek Sharma\\
Cheng Shen\\
Piotr Maj\\
Yasutaka Tanaka\\
You-Cyuan Jhang\\
Vivian Ma\\
Tommi Vehvilainen\\
Kelvin Zou\\
Jeff Nichols\\
Matthew Lei\\
David Qiu\\
Yihao Qian\\
Gokul Santhanam\\
Wentao Wu\\
Yena Han\\
Dominik Moritz\\
Haijing Fu\\
Mingze Xu\\
Vivek Rathod\\
Jian Liu\\
Louis D'hauwe\\
Qin Ba\\
Haitian Sun\\
Haoran Yan\\
Philipp Dufter\\
Anh Nguyen\\
Yihao Feng\\
Emma Wang\\
Keyu He\\
Rahul Nair\\
Sanskruti Shah\\
Jiarui Lu\\
Patrick Sonnenberg\\
Jeremy Warner\\
Yuanzhi Li\\
Bowen Pan\\
Ziyi Zhong\\
Joe Zhou\\
Sam Davarnia\\
Olli Saarikivi\\
Irina Belousova\\
Rachel Burger\\
Shang-Chen Wu\\
Di Feng\\
Bas Straathof\\
James Chou\\
Yuanyang Zhang\\
Marco Zuliani\\
Eduardo Jimenez\\
Abhishek Sundararajan\\
Xianzhi Du\\
Chang Lan\\
Nilesh Shahdadpuri\\
Peter Grasch\\
Sergiu Sima\\
Josh Newnham\\
Varsha Paidi\\
Jianyu Wang\\
Kaelen Haag\\
Alex Braunstein\\
Daniele Molinari\\
Richard Wei\\
Brenda Yang\\
Nicholas Lusskin\\
Joanna Arreaza-Taylor\\
Meng Cao\\
Nicholas Seidl\\
Simon Wang\\
Jiaming Hu\\
Yiping Ma\\
Mengyu Li\\
Kieran Liu\\
Hang Su\\
Sachin Ravi\\
Chong Wang\\
Xin Wang\\
Kevin Smith\\
Haoxuan You\\
Binazir Karimzadeh\\
Rui Li\\
Jinhao Lei\\
Wei Fang\\
Alec Doane\\
Sam Wiseman\\
Ismael Fernandez\\
Jane Li\\
Andrew Hansen\\
Javier Movellan\\
Christopher Neubauer\\
Hanzhi Zhou\\
Chris Chaney\\
Nazir Kamaldin\\
Valentin Wolf\\
Fernando Bermúdez-Medina\\
Joris Pelemans\\
Peter Fu\\
Howard Xing\\
Xiang Kong\\
Wayne Shan\\
Gabriel Jacoby-Cooper\\
Dongcai Shen\\
Tom Gunter\\
Guillaume Seguin\\
Fangping Shi\\
Shiyu Li\\
Yang Xu\\
Areeba Kamal\\
Dan Masi\\
Saptarshi Guha\\
Qi Zhu\\
Jenna Thibodeau\\
Changyuan Zhang\\
Rebecca Callahan\\
Charles Maalouf\\
Wilson Tsao\\
Boyue Li\\
Qingqing Cao\\
Naomy Sabo\\
Cheng Leong\\
Yi Wang\\
Anupama Mann Anupama\\
Colorado Reed\\
Kenneth Jung\\
Zhifeng Chen\\
Mohana Prasad Sathya Moorthy\\
Yifei He\\
Erik Hornberger\\
Devi Krishna\\
Senyu Tong\\
Michael (Taoyi) Lee\\
David Haldimann\\
Yang Zhao\\
Bowen Zhang\\
Chang Gao\\
Chris Bartels\\
Sushma Rao\\
Nathalie Tran\\
Simon Lehnerer\\
Co Giang\\
Patrick Dong\\
Junting Pan\\
Biyao Wang\\
Dongxu Li\\
Mehrdad Farajtabar\\
Dongseong Hwang\\
Grace Duanmu\\
Eshan Verma\\
Sujeeth Reddy\\
Qi Shan\\
Hongbin Gao\\
Nan Du\\
Pragnya Sridhar\\
Forrest Huang\\
Yingbo Wang\\
Nikhil Bhendawade\\
Diane Zhu\\
Sai Aitharaju\\
Fred Hohman\\
Lauren Gardiner\\
Chung-Cheng Chiu\\
Yinfei Yang\\
Alper Kokmen\\
Frank Chu\\
Ke Ye\\
Kaan Elgin\\
Oron Levy\\
John Park\\
Donald Zhang\\
Eldon Schoop\\
Nina Wenzel\\
Michael Booker\\
Hyunjik Kim\\
Chinguun Erdenebileg\\
Nan Dun\\
Eric Liang Yang\\
Priyal Chhatrapati\\
Vishaal Mahtani\\
Haiming Gang\\
Kohen Chia\\
Deepa Seshadri\\
Donghan Yu\\
Yan Meng\\
Kelsey Peterson\\
Zhen Yang\\
Yongqiang Wang\\
Carina Peng\\
Doug Kang\\
Anuva Agarwal\\
Albert Antony\\
Juan Lao Tebar\\
Albin Madappally Jose\\
Regan Poston\\
Andy De Wang\\
Gerard Casamayor\\
Elmira Amirloo\\
Violet Yao\\
Wojciech Kryscinski\\
Kun Duan\\
Lezhi Li\\

\end{multicols}

\end{document}